# The Effects of JPEG and JPEG2000 Compression on Attacks using Adversarial Examples

Ayşe Elvan Aydemir, Alptekin Temizel, Tuğba Taşkaya Temizel
Graduate School of Informatics
Middle East Technical University, Ankara-Turkey
{elvan.gunduz, atemizel, ttemizel}@metu.edu.tr

*Abstract*—Adversarial examples are known to have a negative effect on the performance of classifiers which have otherwise good performance on undisturbed images. These examples are generated by adding non-random noise to the testing samples in order to make classifier misclassify the given data. Adversarial attacks use these intentionally generated examples and they pose a security risk to the machine learning based systems. To be immune to such attacks, it is desirable to have a pre-processing mechanism which removes these effects causing misclassification while keeping the content of the image. JPEG and JPEG2000 are well known image compression techniques which suppress the high-frequency content taking the human visual system into account. JPEG has been also shown to be an effective method in reducing adversarial noise. In this paper, we propose applying JPEG2000 compression as an alternative and systematically compare the classification performance of adversarial images compressed using JPEG and JPEG2000 at different target PSNR values and maximum compression levels. Our experiments show that JPEG2000 is more effective in reducing adversarial noise as it allows higher compression rates with less distortion and it does not introduce blocking artifacts.

## I. INTRODUCTION

Image classification has several practical applications and has been a major interest area for computer vision researchers. Proving their success in ImageNet competitions, deep learning methods have now become the de-facto tools in solving image classification problems. However, recent studies have shown that these methods are vulnerable to malicious attacks. Researchers have revealed that the classifiers could be fooled to misclassify a given image by introducing specially crafted noise to the data (adversarial examples) [1]. Adversarial images having perturbed data points have been shown to make high-performing image classifiers such as Inception [2] misclassify with great confidence. These adversarial examples then have been replicated in further studies [3] [4] [5] [6] and new methods of creating adversarial examples were proposed [5] [7] [8] [9]. As a general principle, adversarial examples are generated by adding non-random noise into the image. The amount of this noise is small enough to be almost imperceptible to human eye. Example adversarial images, generated using the method in [5], are given in Figure 1.

Making the classifiers robust to these adversarial examples is a hot topic of research in the field of image classification. "Defense against Adversarial Attacks" challenge was held in the conference Advances in Neural Information Processing Systems 2017 (NIPS 2017) [10]. The proposed defense mechanisms against *adversarial attacks* include defensive distillation [11], denoising autoencoders [12], compression based approaches [13] and retraining the networks with adversarial examples. It has been shown that JPEG compressing an adversarial image before feeding it to the classification network works as a defense [13] and the classification performance of the OverFeat network improves with the usage of a compression layer.

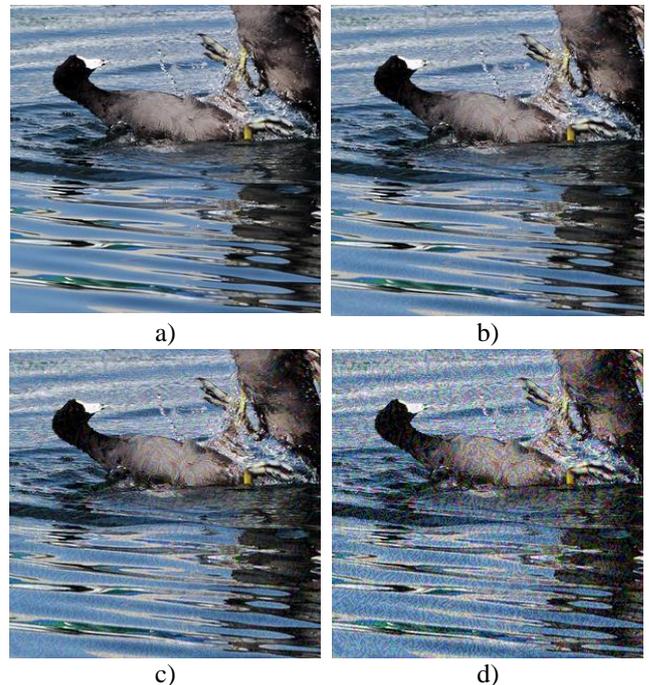

*Figure 1 a) Original image and adversarial images with various ε values: b) ε = 5, c) ε = 10, d) ε = 20*

JPEG is essentially a lossy image compression algorithm exploiting coding redundancy, inter-pixel redundancy and psycho-visual redundancy. *Coding redundancy* is based on the observation that some pixels are more common; Huffman coding assigns shorter bits strings to more frequent symbols. *Inter-pixel redundancy* occurs due to the similarity of neighboring pixels; run-length coding (RLC) exploits this observation and RLC is utilized for a more efficient representation of pixel values. JPEG is a transform based compression algorithm where Discrete Cosine Transform (DCT) is used to transform image blocks into their frequency components. The image is first subdivided into 8x8 blocks and each block is DCT transformed separately. After the transform, the coefficients in the block are quantized using 8x8 quantization tables. The quantization table uses the fact that the coefficients represent increasing frequencies when they are traversed in a zig-zag order. *Psycho-visual redundancy* is exploited by penalizing the higher frequency coefficients more as human visual system is less sensitive to these components. A reflection on these steps reveals that the only step removing any information from the image (disregarding the color space conversion) is the quantization part which suppresses the higher frequencies. While useful as an adversarial defense mechanism, JPEG has two major disadvantages in this context: (*i*) it works by transforming and quantizing 8x8 blocks individually, which inevitably results in blocking artifacts (Figure 2). As these artifacts occur at the same locations for each image, it is expected to reduce the classification performance. (*ii*) JPEG is not designed for high-compression rates and the image is distorted significantly at high rates, adversely affecting the classification performance.

JPEG2000 compression, which is based on wavelet transform, was developed to alleviate the shortcomings of JPEG. Wavelet coding typically produces more efficient compression than DCT based methods. In addition, it does not introduce any blocking artifacts as there is no need to subdivide the image before transformation. Embedded Block Coding with Optimized Truncation (EBCOT) produces a stream having quality and resolution scalability.

It has recently been shown that JPEG compression is effective against adversarial attacks and classification accuracy on adversarial images is recovered when compression is increased. However the accuracy drops when quantization becomes too aggressive [14]. On the other hand, JPEG2000 could achieve higher compression rates with less distortion (i.e. lower mean square error at the same file size) and does not introduce blocking artifacts. These properties are expected to lead to a better performance while removing the adversarial noise.

In this study, we propose using JPEG2000 as an alternative and systematically examine the effects of JPEG and JPEG2000 compressing adversarial images before the classification and compare their performance.

## II. METHOD

There are two main types of adversarial attacks: Targeted attacks and Non-Targeted attacks. Targeted attacks are types of attacks that pre-determine the target incorrect class for each image. For instance, if the aim is to make the classifier incorrectly classify a cat image as a dog image, the target would be the "dog" class. Some examples for targeted attacks are Momentum Iterative Attack [15], Saliency Map Method [16] and Carlini-Wagner attack [4]. Non-targeted attacks are types of attacks which do not explicitly specify the target outcome. In both cases, the maximum amount of noise that can be added to image is determined using an epsilon value. In this work, we used non-targeted attacks to test our ideas.

To test the efficacy of JPEG and JPEG2000 compression, we added adversarial noise to images in [17] using the Fast Gradient Sign (FGSM) method introduced in [5] as well as Basic Iterative Method introduced in [8]. In the experiments, we used Inception v3 classifier [2] and the widely known ImageNet Dataset [17]. As we have used pre-trained models, we accessed only the test set to perform classification. The test set has 1000 annotated images.

FGSM attack applies noise linearly based on the gradient of the objective function while BIM attacks applies FGSM with small increments.

The attacks were applied with varying ε values. ε values determine the maximum amount of adversarial noise that can be added to the image to fool the classifier. In our case, we used 5, 10 and 20 for FGSM attack while we kept the ε value constant at 15 for BIM attack. The ε for BIM attack was kept constant because BIM attack takes longer to run due to its iterative nature. These values are also coherent with the study that we are elaborating on [13].

Adversarial images were compressed using JPEG and JPEG2000 with different settings. To test the performance of both compression schemes, JPEG and JPEG2000 compression were applied to the images in a fashion that the resulting image would have a target PSNR value. For completeness, we tested with various PSNR values and reported the results in Section III. The PSNR values used for the compression are 23, 25, 28 and 31 dB. These values are chosen arbitrarily by visually judging the image quality. The sample images for these compression values are given in Figure 2.

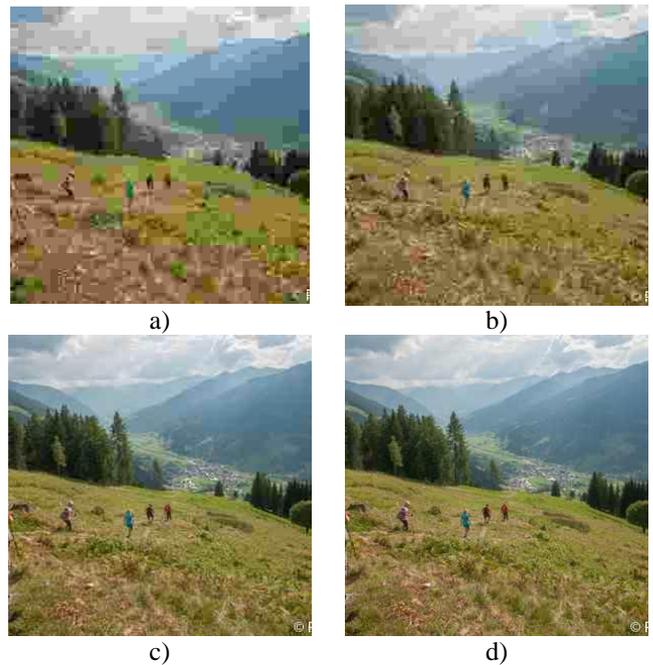

*Figure 2 – JPEG Compressed Images at Different Quality Levels. Blocking artifacts can be observed with increasing compression rate: a) 23 dB PSNR b) 25 dB PSNR c) 28 dB PSNR d) 31 dB PSNR*

PSNR based compression is inherent in JPEG2000 due to the EBCOT process, however it is not inherent to JPEG compression. In order to achieve comparable PSNR to preset values, JPEG compression was applied iteratively until the resulting image's PSNR is within 0.01 difference to the target value.

To further explore the limits of these two compression schemes, we reduced the images to the smallest possible size that our tools allowed us to. This case is aimed to study the change in performance when maximum amount of suppression was applied. The smallest possible size for JPEG2000 was found to be around 5 KB, but for JPEG compression this value varied significantly ranging from 9 KBs to 11.5 KBs for different noise levels.

## III. RESULTS

To assess the classification performance, the test set of ImageNet is used. This test set contains 1000 images previously not seen by the classifier. The accuracy of the classifier is reported for each different case along with the accuracy for uncompressed adversarial examples.

Accuracy values for PSNR based analysis can be found in Table I. The columns refer to the attacks along with the ε value used for the attack. The rows refer to different compression settings and PSNR values are calculated using the original uncompressed adversarial image. The accuracy for the original adversarial attack image is given in the last row. The highest accuracy values for each column are given in bold.

*Table 1 Accuracy results for images compressed at different PSNR values*

|  | PSNR (dB) | FGSM (ε =20) | FGSM (ε =10) | FGSM (ε =5) | BIM (ε =15) |
|---|---|---|---|---|---|
| JPEG | 23 | **0.361** | 0.415 | 0.379 | 0.393 |
|  | 25 | 0.317 | 0.457 | 0.499 | 0.452 |
|  | 28 | 0.283 | 0.319 | 0.502 | 0.229 |
|  | 31 | 0.283 | 0.260 | 0.369 | 0.031 |
| JPEG2000 | 23 | 0.339 | **0.473** | 0.513 | **0.461** |
|  | 25 | 0.337 | 0.436 | **0.577** | 0.429 |
|  | 28 | 0.283 | 0.346 | 0.490 | 0.164 |
|  | 31 | 0.281 | 0.265 | 0.378 | 0.045 |
| Uncompressed | NA | 0.266 | 0.244 | 0.221 | 0.016 |

Accuracy values for maximum compression are given in Table 2. As the images are compressed to their minimum size, there is no PSNR setting for this case.

*Table 2 Accuracy results at maximum compression levels*

|  | FGSM (ε =20) | FGSM (ε =10) | FGSM (ε =5) | BIM (ε =15) |
|---|---|---|---|---|
| JPEG2000 | **0.428** | **0.523** | **0.634** | **0.511** |
| JPEG | 0.378 | 0.400 | 0.475 | 0.229 |
| Uncompressed | 0.266 | 0.244 | 0.221 | 0.016 |

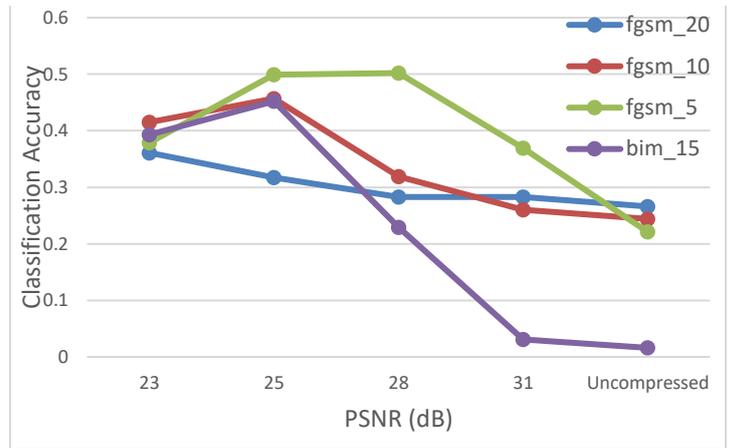

*Figure 3 Accuracy for JPEG for different PSNR values*

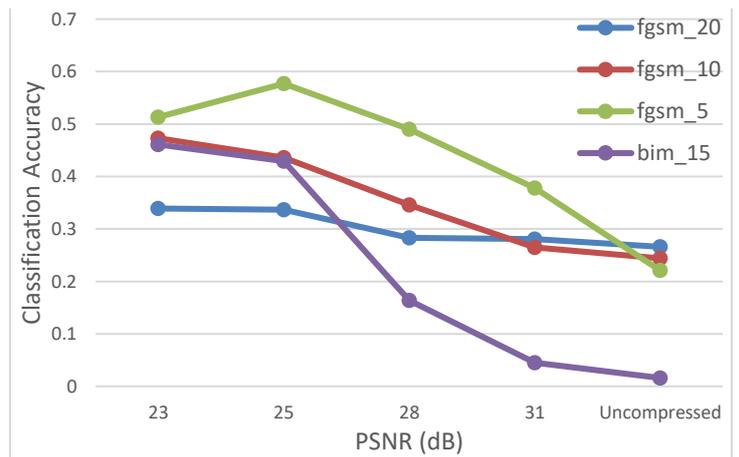

*Figure 4 Accuracy for JPEG2000 for different PSNR values*

## IV. DISCUSSION

The purpose of this study was to explore the effectiveness of using compression algorithms for pre-processing before feeding these images to the classifiers. The classification results were improved by a significant margin using this approach which suggests that suppressing high-frequency content is an effective defense mechanism against adversarial attacks. We used the compression algorithms as a whole to test this idea in line with the previous works [13] and we changed the compression levels to observe its effects. However, it is evident that using only the transformation and quantization parts should be sufficient and computationally more efficient to achieve similar results.

JPEG compression uses 8x8 filter windows and as a result adds artificial block borders to the compressed image (Figure 5). As the block partitioning is fixed, the blocking artifacts occur at the same pixel positions for each image. This is expected to make images more similar at these borders and reduce the classification performance. While the classification performance of JPEG2000 compression is noticeably better than JPEG in case of maximum compression, this cannot be claimed for PSNR based analysis where JPEG compression outperforms JPEG2000 compression.

We have observed the best performance when the images are compressed to minimum possible size. There was a significant range in image sizes and adversarial noise reduction when we used PSNR based compression. This lead us to think, adding adversarial noise to images, and then applying PSNR based compression sometimes fails to compress the image enough to reduce high-frequency noise as it tries to preserve adversarial noise as well. So, maximum compression performs better as it suppresses the higher-frequency components more.

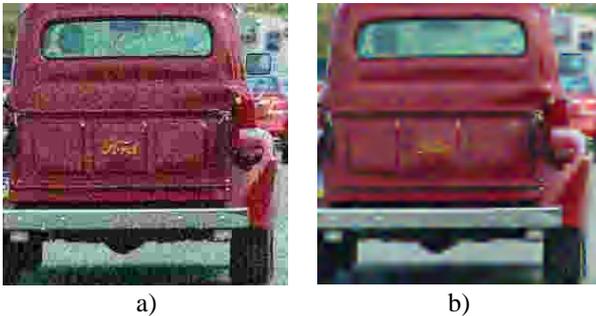

*Figure 5 a) JPEG compressed b) JPEG 2000 compressed image*

## V. CONCLUSION

In this paper, we explored the potential of JPEG2000 compression as defense against adversarial examples and compared its performance with JPEG. The results show that while both increase the classification accuracy of adversarial images and work effectively against Fast Gradient Sign Method and Basic Iterative Method, JPEG2000 performs better and outperform JPEG especially higher compression rates. A closer look into the workings of these compression methods reveals that suppression of high-frequency content is the main factor for the performance gain. In the future, we are planning to use transformations (DCT and wavelet) followed by quantization as preprocessing steps instead of using the whole compression algorithm as a black box. This will also enable a more controlled use of different quantization matrices and by representing the quantization matrix parametrically would also allow learning these parameters for a particular defense.

REFERENCES

[1] C. Szegedy, W. Zaremba, I. Sutskever, J. Bruna, D. Erhan, I. Goodfellow and R. Fergus, "Intriguing properties of neural networks," 2014. [Online]. Available: https://arxiv.org/abs/1312.6199. [Accessed 3 2 2018].

[2] C. Szegedy, V. Vanhoucke, S. Ioffe and J. Shlens, "Rethinking the Inception Architecture for Computer Vision," in *The IEEE Conference on Computer Vision and Pattern Recognition (CVPR)*, 2016.

[3] O. Bastani, Y. Ioannou, L. Lampropoulos and D. Vytiniotis, "Measuring neural net robustness with constraints," 2016. [Online]. Available: https://arxiv.org/pdf/1605.07262.pdf. [Accessed 3 2 2018].

[4] N. Carlini and D. Wagner, "Towards Evaluating the Robustness of Neural Networks," 2017. [Online]. Available: https://arxiv.org/abs/1608.04644. [Accessed 3 2 2018].

[5] I. J. Goodfellow, J. Shlens and C. Szegedy, "Explaining and Harnessing Adversarial Examples," in *arXiv preprint arXiv:1412.6572*, 2014.

[6] R. Huang, B. Xu, D. Schuurmans and C. Szepesvari´, "Learning with a strong adversary," 2015. [Online]. Available: https://arxiv.org/pdf/1511.03034.pdf. [Accessed 3 2 2018].

[7] S.-M. Moosavi-Dezfooli, A. Fawzi and F. P., "DeepFool: Simple and Accurate Method to Fool Deep Neural Networks," in *IEEE Conference on Computer Vision and Pattern Recognition (CVPR)*, 2016.

[8] A. Kurakin, I. J. Goodfellow and S. Bengio, "Adversarial Examples in the Physical World," in *ICLR*, 2017.

[9] T. Miyato, S.-c. Maeda, M. Koyama, K. Nakae and S. Ishii, "Distributional Smoothing With Virtual Adversarial Training," in *ICLR*, 2016.

[10] NIPS 2017; Kaggle, "NIPS 2017: Defense Against Adversarial Attacks," Kaggle, NIPS2017, [Online]. Available: https://www.kaggle.com/c/nips-2017-defense-against-adversarial-attack. [Accessed 3 2 2018].

[11] N. Papernot, P. McDaniel, X. Wu, S. Jha and A. Swami, "Distillation as a Defense to Adversarial Perturbations against Deep," 2016. [Online]. Available: https://arxiv.org/abs/1511.04508. [Accessed 3 2 2018].

[12] S. Gu and L. Rigazio, "Towards deep neural network architectures robust to adversarial examples," 2014. [Online]. Available: https://arxiv.org/pdf/1412.5068.pdf. [Accessed 3 2 2018].

[13] G. K. Dziugaite, Z. Ghahramani and D. M. Roy, "A study of the effect of JPG compression on adversarial images," 2016.

[14] N. Akhtar and A. Mian, "Threat of Adversarial Attacks on Deep Learning in Computer Vision: A Survey," 2018. [Online]. Available: arXiv preprint arXiv:1801.00553, 2018..

[15] Y. Dong, F. Liao, T. Pang and X. Hu, "Boosting Adversarial Attacks with Momentum," 5 Dec 2017. [Online]. Available: https://arxiv.org/pdf/1710.06081.pdf. [Accessed 3 2 2018].

[16] N. Papernot, P. McDaniel and S. Jha, "The Limitations of Deep Learning in Adversarial Settings," in *IEEE European Symposium on Security & Privacy*, 2016.

[17] J. Deng, W. Dong, R. Socher, L.-J. Li, K. Li and L. Fei-Fei, "ImageNet: A Large-Scale Hierarchical Image Database," in *CVPR09*, 2009.